\pdfoutput=1

\documentclass[11pt]{article}

\usepackage[final]{acl}

\usepackage{times}
\usepackage{latexsym}

\usepackage[T1]{fontenc}

\usepackage[utf8]{inputenc}

\usepackage{microtype}

\usepackage{inconsolata}

\usepackage{booktabs}
\usepackage{graphicx}
\usepackage{multirow}
\usepackage{subcaption}

\usepackage{enumitem}
\setlist{nolistsep}

\title{STAIR (STructure Aware Information Retriever): A novel dataset and LLM based retriever for document structure augmentation}
\author{
  \normalsize\textbf{Vineet Kumar\footnotemark[1] \quad Meghanadh Pulivarthi \quad Vishwajeet Kumar \quad Jaydeep Sen} \\
  \normalsize\textbf{Riyaz Ahmad Bhat \quad Sachindra Joshi} \\
  \normalsize IBM \\
  \normalsize\texttt{vineet.mundhra@gmail.com \quad vishk024@in.ibm.com} \\
  \normalsize\texttt{jaydesen@in.ibm.com \quad Riyaz.Bhat@ibm.com \quad jsachind@in.ibm.com}
}

\newcommand{\sys}{{\textbf{STAIR}}}
\newcommand{\ben}{{\textbf{SearchTome}}}

\newcommand{\basellm}{{Mistral Instruct v0.2}}
\newcommand{\devmetric}{{Recall@1}}
\newcommand{\toc}{\textbf{{ToC}}}
\newcommand{\eat}[1]{}

\renewcommand{\thefootnote}{\fnsymbol{footnote}}

\begin{document}
\maketitle

\footnotetext[1]{Work done while at IBM. Currently at Amazon Books Science.}
\renewcommand{\thefootnote}{\arabic{footnote}}  

\begin{abstract}
Retrieval Augmented Generation (RAG) is a key component for generating accurate and hallucination free answers using Large Language Models (LLMs). LLMs are improving at handling long context, but still suffer from ``lost in the middle'' problem. Thus, precise and accurate retrieval is important. Current retrievers chunk long context into length-based manageable chunks -- in the process throwing away rich and informative semantic global structure in the corpus. We introduce a novel retrieval system {\sys} that empowers an LLM to exploit global structure in a corpus such as a Table of Contents ({\toc}) to efficiently store and retrieve information from its model parameters. Our thorough and careful ablation studies with a finetuned Differentiable Search Index (DSI) system show that {\toc} helps build a low hallucination (less than 0.05\%) generative Information Retrieval (IR) system and can generalize to examples where very few training samples are available. To further research in this novel direction of {\toc} based retrieval we release {\ben} -- a diverse benchmark created from 18 books across 6 diverse domains to further research in this novel direction. {\sys} achieves a high Recall@1 score of 82.6\% on {\ben} as compared to DSI (76.9\%), where the difference is found to be statistically significant. {\sys} easily beats other strong baselines such as BM25 (59.5\%), DPR (68.7\%) and out-of-the-box Mistral (13.8\%). The code and benchmark dataset for {\sys} are publicly available on \href{https://github.com/meghanadhpulivarthi/stair-clean.git}{GitHub}
and \href{https://huggingface.co/datasets/meghanadh/SearchTome}{Hugging Face},
respectively.

\end{abstract}

\section{Introduction}
\label{intro}
The burgeoning interest in Retrieval Augmented Generation (RAG) has led to a significant surge in the development of advanced Information Retrieval (IR) systems. Large Language Models (LLMs) in turn can now handle large contexts~\cite{chen2023extending,liu2024world}, though they suffer from a ``lost in the middle'' problem ~\cite{liu-etal-2024-lost,bai2024longbench,bai2024longbench2, Li2024LongcontextLS}. Therefore, retrieving precise information~\cite{Pipitone2024LegalBenchRAGAB} is extremely important to curb hallucinations~\cite{Laban2024SummaryOA} and generate accurate responses. Current retrievers address this by creating length based chunks~\cite{Setty2024ImprovingRF} and throwing away rich and informative semantic global structure in the corpus. This leads to a sub-optimal retrieval quality -- length-based chunks compete with each other due to a lack of semantic coherence and boundaries.

In this work, we address this key limitation by augmenting the retriever with a structured global view of the corpus. Global structured view over a long context helps knowledge ingestion~\cite{liu2024structureaware}. Further, LLMs are capable of storing the entire corpus in its model parameters to directly generate a document identifier for a user query~\cite{Tay2022TransformerMA}. We posit that by empowering an LLM with a global structure of the search corpus, it can store and retrieve information more accurately from its model parameters. Such a global structure already exists for a Wikipedia~\cite{wikipedia-sample-layout} page, textbooks, and enterprise help and product feature webpages~\cite{SAP-product-documentation} and technical reports~\cite{SEC-10k-structure}.

\begin{figure*}[!htb]
\begin{center}
    \includegraphics[width=\linewidth]{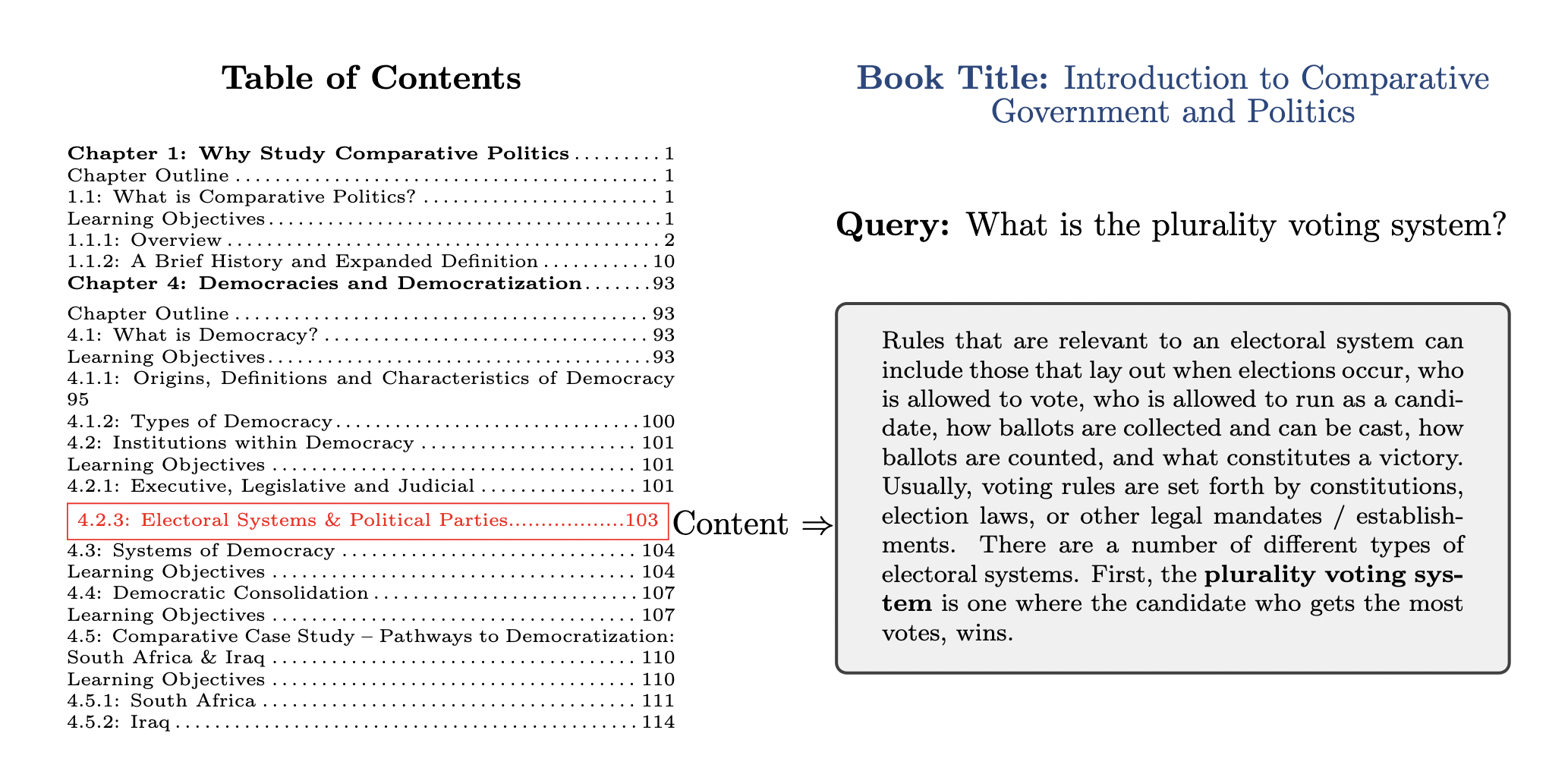}
    \caption{A human can answer ``What is the plurality voting system?'' by looking at the table of contents and picking Section 4.2.3 without even reading the book \href{https://open.umn.edu/opentextbooks/textbooks/1435}{Introduction to Comparative Government and Politics}.}
    \label{fig:intro_example}
\end{center}
\end{figure*}

Figure~\ref{fig:intro_example} demonstrates that for a question such as ``What is the plurality voting system?'', it is  natural for a human to consult the {\toc} and first narrow down that it could be answered by the Chapter ``Democracies and Democratization''. Further subsections reveal that perhaps the Section ``Institutions within Democracy'' and the subsection ``Electoral Systems \& Political Parties'' could contain the answer. Content for this subsection indeed contains the definition for the plural voting system. Text content associated with {\toc} entries is semantically coherent organized around a topic as captured by the title of the {\toc} entry. Further, a {\toc} entry has a  clear semantic boundary with other {\toc} entries. {\toc} as a unit of retrieval is a novel yet intuitive way of organizing long context for retrieval.


In this work, we present {\sys} (\textbf{ST}ructure \textbf{A}ware \textbf{I}nformation \textbf{R}etriever) for {\toc} based retrieval, inspired by how a human searches for information using {\toc}, and the fact that LLMs can store the entire corpus in the model's parameters. Our careful and thorough ablation studies reveal that empowering an LLM with {\toc} helps in two critical dimensions - curbing hallucinations and generalizing for content with a low number of training examples. {\sys} achieves a high Recall@1 score of 82.6, which is is ($+7.4\%$) better than its closest baseline Differentiable Search Index (DSI)~\cite{Tay2022TransformerMA}. DSI needs to discover the book structure (indirectly) through the completion part of training examples, while {\sys}\ can use exploit the structure provided as part of the input to accurately store and retrieve information from its model's parameters.


We develop and release a new multi-domain benchmark {\ben}, made up of $18$ books of varying sizes across $6$ domains, parsed and cleaned to extract their {\toc} mapped to the text of the pertinent sections. We also release a set of train, dev and test queries for each book as part of the benchmark, with gold {\toc} entries labeled for each query serving as the retrieval output. We believe this will be a first step towards promoting further research on {\toc} based retrieval, with {\ben} serving as a standard evaluation benchmark. We train and evaluate {\sys} using {\ben} and compare it with strong baseline retrievers like BM25 and a fully fine-tuned Sequence-to-Sequence model, such as DSI~\cite{Tay2022TransformerMA}, which does not utilize {\toc}. Our experimental results show that {\sys} significantly benefits from incorporating {\toc} in its training and easily outperforms the traditional DSI. Our main contributions are as follows:
\begin{itemize}
\item We propose a novel direction leveraging the Table of Contents ({\toc}) of a lengthy corpus, like a book, as a valuable element for efficiently indexing corpus knowledge and utilizing {\toc} entries as effective retrieval units.
\item We build and release a multi-domain and diverse benchmark {\ben}, consisting of $18$ books from $6$ diverse domains for the task of {\toc} retrieval
\item We develop an LLM based {\sys} and evaluate it on {\ben} and compare it with strong competitive baselines
\end{itemize}

\section{Related Work}
\label{sec:relatedwork}
Efficient and precise retrieval is a critical research problem in natural language processing, with a wide range of applications such as semantic search~\cite{semantic_search}, open domain question answering~\cite{chen-yih-2020-open,zhu2021retrieving} and Retrieval Augmented Generation(RAG)~\cite{lewis2020retrieval} to name a few.
\eat{There have been significant advancements in retriever paradigms, evolving from traditional lexical retrievers like BM25~\cite{robertson2009probabilistic}, which utilizes sparse inverted indexes, to modern dense embedding-based retrievers such as DPR~\cite{Karpukhin2020DensePR} and ColBERT~\cite{Khattab2020ColBERTEA}, which rely on dedicated VectorDBs~\cite{johnson2019billion} for indexing and retrieval. A more recent approach is Differential Search Indexing (DSI)\cite{Tay2022TransformerMA}, which introduces a sequence-to-sequence model that maps queries directly to document identifiers, thereby embedding corpus knowledge within the model parameters.} With the rise of RAG applications, precise retrieval has become increasingly important, particularly in curbing  hallucination in generative models. \eat{Consequently, a central research focus across all retriever paradigms is the development of more effective indexing strategies to enhance retrieval quality} We can broadly classify all retrieval techniques into three main categories:

\textbf{Dense Retrieval}~\cite{Cai2021SemanticMF,Karpukhin2020DensePR} represent a query and document using a dense vector and compute the similarity based on the distance between their vectors. DPR~\cite{Karpukhin2020DensePR} uses dual encoder, ColBERT ~\cite{Khattab2020ColBERTEA} represents every token in the query and document using dense vector and performs late interaction to score query and document pairs. Recently,~\citet{sarthi2024raptor} introduced RAPTOR, a method for building a hierarchical tree by recursively embedding, clustering, and summarizing content from lengthy documents to retrieve relevant text at various levels of abstraction. Our approach differs in two key aspects: while RAPTOR uses dense retrievers for indexing and retrieval, we adopt model-based indexing. Additionally, we enhance model-based indexing by leveraging table-of-contents ({\toc}) structures, unlike RAPTOR, which relies solely on dense retrievers to access information at multiple levels of abstraction.

\textbf{Learned Sparse Retrieval} systems use sparse vector representation for a query and document. Unlike traditional sparse retrieval methods like BM25\cite{robertson2009probabilistic}, which rely on exact token matches, these models learn to represent queries and documents in a sparse high-dimensional space, allowing for efficient lexical matching. SPLADE~\cite{Hai2023CoSPLADECS} (Sparse Lexical and Expansion Model for Information Retrieval) is a prominent example of such systems. Several other models have been developed to enhance sparse retrieval, including SPLADEv2 \cite{formal2021splade}, DeepCT\cite{dai2019context}, uniCOIL\cite{lin2021few} and DeepImpact~\cite{basnet2024deeperimpact}.

\textbf{Model-based Indexing}~\cite{Metzler2021RethinkingSM} such as Differential Search Index (DSI)~\cite{Tay2022TransformerMA} embed knowledge of the entire corpus directly into its model parameters -- greatly simplifying the retrieval process. {\sys} takes this paradigm one step further by learning to exploit a rich hierarchal and semantic global structure within the corpus.

\section{{\ben}: A new benchmark for Table of Contents based retrieval}
\label{sec:benchmark}
\begin{table*}[h!]
\small
\caption{{\ben} Book links \& Statistics}
\label{table:benchmark}
\resizebox{\textwidth}{!}{%
\begin{tabular}{llllllll}
\textbf{Domain} & \textbf{Book} & \textbf{\#Pages} & \textbf{\#Leaves} &\textbf{Test}&\textbf{Train}&\textbf{Dev}\\
\toprule
\multirow{3}{*}{Education} & \href{https://open.umn.edu/opentextbooks/textbooks/1254}{Open Music Theory} & 1297 & 429 & 7430 & 4493 & 1671  \\
& \href{https://open.umn.edu/opentextbooks/textbooks/1621}{The Whole Child: Development in the Early Years} & 182 & 129 & 1746 & 1056& 388 \\
&  \href{https://open.umn.edu/opentextbooks/textbooks/221}{Teaching in a Digital Age} & 779 & 118 & 6150&3797&1489\\
\midrule
\multirow{3}{*}{Finance} & \href{https://open.umn.edu/opentextbooks/textbooks/principles-of-financial-accounting}{Principles of Financial Accounting} & 318 & 98 & 4119&2395&981 \\
& \href{https://open.umn.edu/opentextbooks/textbooks/financial-accounting}{Accounting in the Finance World} & 572 & 80 & 4481&2747&1081 \\
& \href{https://open.umn.edu/opentextbooks/textbooks/financial-and-managerial-accounting}{Financial and Managerial Accounting} & 1077 & 107 & 7815&4692&1901\\
\midrule
\multirow{3}{*}{Law} & \href{https://open.umn.edu/opentextbooks/textbooks/1219}{Construction Contracting} & 403 & 103 & 3653&2249&866\\
& \href{https://open.umn.edu/opentextbooks/textbooks/criminal-procedure-undergraduate-edition-author}{Criminal Procedure} & 897 & 106 & 5610& 3401& 1349 \\
& \href{https://open.umn.edu/opentextbooks/textbooks/tort-law-cases-and-commentaries}{Tort Law: Cases and Commentaries} & 948 & 396 & 18209&11174&4363 \\
\midrule
\multirow{3}{*}{Medicine} & \href{https://open.umn.edu/opentextbooks/textbooks/1321}{Nursing Assistant} & 659 & 118 & 4586&2790&1093 \\
& \href{https://open.umn.edu/opentextbooks/textbooks/1013}{Nursing Fundamentals} & 1327 & 121 & 9065&5493&2200 \\
& \href{https://open.umn.edu/opentextbooks/textbooks/1427}{Nursing Management and Professional Concepts} & 599 & 74 & 3695&2277&885 \\
\midrule
\multirow{3}{*}{Natural Sciences} & \href{https://open.umn.edu/opentextbooks/textbooks/1279}{Introduction to Genetics} & 513 & 70 & 2254&1345&527 \\
& \href{https://open.umn.edu/opentextbooks/textbooks/1600}{Principles of Mechanics} & 179 & 125 & 2341&1418&543 \\
& \href{https://open.umn.edu/opentextbooks/textbooks/organic-chemistry}{Organic Chemistry} & 1249 & 321 & 11124&6601&2635 \\
\midrule
\multirow{3}{*}{Social Sciences} & \href{https://open.umn.edu/opentextbooks/textbooks/1446}{Foundations of Aural Skills} & 674 & 118 & 2451&1421&559 \\
& \href{https://open.umn.edu/opentextbooks/textbooks/1435}{Introduction to Comparative Government and Politics} & 421 & 189 & 3893&2288&898 \\
& \href{https://open.umn.edu/opentextbooks/textbooks/a-practicum-in-behavioral-economics}{A Practicum in Behavioral Economics} & 381 & 178 & 4627&2821&1073 \\
\end{tabular}
}
\end{table*}

\subsection{Breaking the Mold: Why a New Benchmark Is Needed}
There are multiple benchmarks proposed around long context applications such as ContractNLI~\citep{koreeda-manning-2021-contractnli-dataset} focused on NLI or \textbf{Scrolls~\citep{shaham-etal-2022-scrolls}} with seven challenging tasks. However, there doesn't exist any benchmark for long context retrieval with any form of structured view, which we hypothesize as the key for precise retrieval. Benchmarks such as GovReport~\citep{huang2021govreport}, SummScreen~\citep{chen-etal-2022-summscreen}, and QM-Sum ~\cite{zhong2021qmsum} focus on generating summaries of a length document, whereas, Qasper ~\citep{dasigi2021qasper}, QuALITY~\citep{pang-etal-2022-quality} and NarrativeQA~\citep{kocisky2018narrativeqa} focus on the generation aspect with \textbf{only answers} for a user query and \textbf{do not contain the gold passage for retrieval}. The closest benchmark is \textbf{LocoV1~\citep{SaadFalcon2024BenchmarkingAB}} which does include gold passages for each query. However, due to \textbf{lack of {\toc} or a global structure for the input document}, we cannot use this benchmark for our task.\\ To the best of our knowledge, {\ben} is the first benchmark which provides a clean structured view with Table of Contents for evaluating long context retrieval with structures . 

\subsection{{\ben}: A diverse and novel benchmark for {\toc} Retrieval}

Our main motivation for a new benchmark is to evaluate how well a global semantic structure such as a {\toc} can help in better retrieval for long context applications which is typical for technical reports or voluminous books. Thus to create a clean and effective benchmark we turn to \href{https://open.umn.edu/opentextbooks/}{opentextbooks}, which is perhaps the largest collection of such textbooks. To make the benchmark a comprehensive one across domains, we picked six diverse domains namely -- Education, Finance, Law, Medicine, Natural Sciences and Social Sciences and selected three books from each domain (Table~\ref{table:benchmark}). We parsed the PDF for each book and extracted Table of Contents using pymupdf\footnote{\url{https://pymupdf.readthedocs.io/en/latest/}}. A sample of the final cleaned content for one of the books can be viewed \href{https://anonymous.4open.science/r/s_331/data/education/book1/v1/content.jsonl}{here}. For each paragraph we ask a powerful LLM Mixtral 8x7b model~\citep{jiang2024mixtral} to generate multiple questions covering all important topics in the paragraph~\cite{Zhang2024SyntheticKI}. Following the same technique as was proposed in DSI~\cite{Tay2022TransformerMA} to train the parametric index, we use a portion of these questions for training DSI~\cite{Tay2022TransformerMA} and {\sys}, a small portion as development set which is to help us pick the best checkpoint. Majority of the generated questions were picked as test questions -- to thoroughly test coverage of content by an IR system. We believe our benchmark {\ben} will further research in building IR systems which can leverage structure in a corpus and retrieve the most relevant chunks for a user query.

\section{{\sys}: Table of Contents Searcher}
\label{sec:system}
We now formally define the problem statement and the technique for retrieving a section title given a book and its Table of Contents ({\toc})

\subsection{Notations}
We are given a long document $D$ and its Table of Contents ${{\toc}}_D = \{T_{1}, T_{2}, ....,T_{n}\}$. An edge $e: T_{p} \rightarrow T_{c}$ between two nodes $T_{p}$, $T_{c}$ $\in {{\toc}}_D$ is defined as $T_{p}$ = parent($T_{c}$), if the section represented by title $T_{p}$ is further divided into multiple sub-sections including $T_{c}$. The set of leaf nodes can thus be defined as $LN_{D}$=$\{T_{l} \in {\toc}_{D} | \exists! T_{c} \in {\toc}_{D}, T_{l}=parent\_of(T_{c})\} \subseteq {{\toc}}_{D}$. Our goal is to retrieve the correct leaf node $T_{l} \in LN_{D}$ whose content can answer a user query $q$.


\subsection{Training {\sys}}
\label{subsec:train_task}
\begin{figure*}[!tbh]
    \centering
    \includegraphics[width=\linewidth]{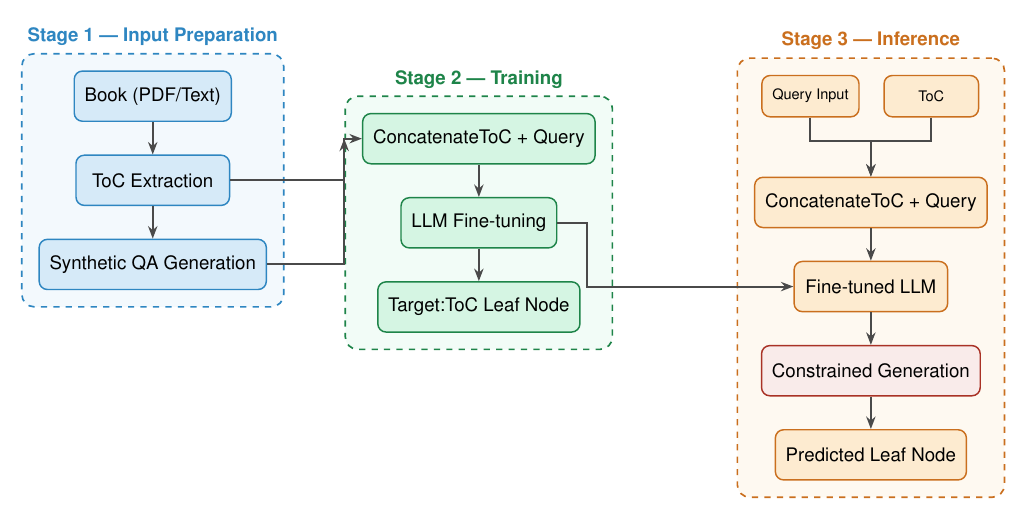}
    \caption{Overview of our pipeline. Stage 1 prepares synthetic QA data
             from the input book. Stage 2 fine-tunes a language model to map
             a ToC-augmented query to the correct leaf node. Stage 3 runs
             inference with constrained generation, restricting the output
             vocabulary to valid ToC leaf nodes.}
    \label{fig:arch}
\end{figure*}
\noindent Figure~\ref{fig:arch} depicts the training pipeline for {\sys} which learns two correlated but \textit{different} tasks:\\
\textbf{Corpus knowledge ingestion:} {\sys} needs to learn the information trove $D$ by mapping a book's text content to the corresponding section titles $T_{i} \in {\toc}_{D}$. This is facilitated by the training data, which consists of a series of queries that are linked to their corresponding section titles. Consequently, this phase of training integrates the book's knowledge into the LLM's parameters, similar to model-based indexing systems like DSI, but with a key difference. Unlike other model-based indexing systems, the parameters in {\sys} remember the corpus knowledge not by a document identifier, but by using fine-grained section title entries in the {\toc}, which are likely to have a much stronger semantic connection with the content being ingested into parametric knowledge.

\textbf{Learning to generate from {\toc}:} {\sys} learns to pick the most relevant leaf node from the complete global structure ({\toc}) of a document.
We use the training split from {\ben} to train {\sys} through supervised fine-tuning. During training, we provide the query $q$ along with the complete ${\toc}_{D}$ as input. Consequently, the complete input during training consists of \textit{prompt} + $(q,{\toc}_{D})$. Note that ${\toc}_{D}$, remains consistent across queries from the same book, while the prompt remains unchanged throughout the training process. We fine-tune the model to generate the correct leaf node $T_{l}\in LN_{D}$ that corresponds to the query $q$.

\begin{table}[!htb]
\small
\caption{Sample training example}
\label{table:train_example_small}
\begin{tabular}{l}
\textbf{Prompt} \\
\toprule
\#\#\# Book: Teaching in a Digital Age\\
Pick the best section from the table of contents below\\
which can answer the user query below. Only generate \\
the section name, do not generate any explanation!\\
\\
\#\#\#\ Table of Contents\\
1 Chapter 1: Fundamental Change in Education \\
$\cdots$ \\
leaf 12.7 Step five:master the technology \\
leaf 12.8 Step six:set appropriate learning goals \\
leaf 12.9 Step seven:design course structure and learning\\
$\cdots$ \\
leaf 13.7 Building the future \\
\#\#\# Query: What is the suggested time allocation \\ 
for students studying a course or program? \\
\#\#\# Section: leaf \\
\\
\textbf{Completion} \\
\midrule
12.9 Step seven: design course structure and learning
\end{tabular}
\end{table}

\section{Experiments}
\label{sec:expt}
We report Recall@1 (R@1), Recall@3 (R@3) and Normalized Discounted Cumulative Gain (nDCG@3) metrics on the test set for each book. BeIR~\cite{thakur2021beir} is used for computing the metrics.



\subsection{Baselines}
\label{sec:baselines}
\textbf{Mistral} (Out of the Box LLM): Beam search is used to generate the Top-$K$ predictions for Mistral Instruct v0.2\\
\textbf{BM25}~\cite{robertson2009probabilistic}: We index content using Elastic Search v 8.11.2. Leaf node acts as a unique document identifier for a book\\
\textbf{DPR}: We use NV-Embed-v2\cite{lee2024nv} out of the box with text lengths set to 512 for passages and 256 for queries. NV-Embed-v2 is the leading text-only embedding model with open weights on the MTEB leaderboard\footnote{As per Jan 23'2025 MTEB Leaderboard Snapshot.}. The model is based on Mistral-7B-v0.1\footnote{https://huggingface.co/mistralai/Mistral-7B-v0.1}, making it a suitable baseline for {\sys} as both are finetuned over Mistral.\\
\textbf{DSI~\citep{Tay2022TransformerMA}} infuses the entire knowledge of a corpus in the parameters of an LLM and directly generates a document identifier. We fine-tune DSI using the train split for each book.

\subsection{Fine-tuning {\sys} and DSI}
\label{sec:finetune}
We fine-tune Mistral Instruct v0.2\footnote{https://huggingface.co/mistralai/Mistral-7B-Instruct-v0.2}~\cite{Jiang2023Mistral7} for a maximum of 200 epochs using a LoRA adapter~\cite{Hu2021LoRALA} with a $r=16$ and $\alpha=32$. We early stop with a patience of 20 epochs, by computing {\devmetric} on the dev set. We set the maximum input length to 14k tokens for {\sys} and $512$ tokens for DSI and a maximum output length of 64 tokens. See Table~\ref{table:train_example_small} for a sample training input and output for {\sys}.





\section{Results and Analysis}
\label{sec:results}
\begin{table*}[ht!]
\small
\caption{Examples of model predictions. Gold leaf is highlighted in \textbf{bold} in Table of Contents}
\label{table:pred_examples}
\resizebox{\textwidth}{!}{%
\begin{tabular}{l|l}
\multicolumn{2}{c}{\textbf{Example 1:}} \\
\toprule
\multicolumn{2}{l}{\textbf{Book}: \href{https://open.umn.edu/opentextbooks/textbooks/1621}{The Whole Child: Development in the Early Years}}\\
\multicolumn{2}{l}{\textbf{Query}: How do preschoolers react when caregivers and teachers belittle their autonomous actions?}\\
\midrule
\textbf{Table of Contents} & \textbf{Content}\\
$\cdots$\\
2 Chapter Two: Theorists and Theories of Development & Initiative vs. Guilt (Preschool Years)\\
leaf 2.1 Theories of Development & The development of courage and independence ...\\
2.2 Psychosocial Theory & ...\\
leaf 2.2.1 Trust vs. Mistrust (Infancy)& ...\\
leaf 2.2.2 Autonomy vs. Shame/Doubt (Toddlerhood) &...\\
\textbf{leaf 2.2.3 Initiative vs. Guilt (Preschool Years)}\\
leaf 2.2.4 Industry vs. Inferiority (Elementary Years) & If caregivers and preschool teachers encourage and support\\
leaf 2.2.5 Identity vs. Role-Confusion (Adolescence) & children’s efforts while also helping them make realistic \\
leaf 2.2.6 Intimacy vs. Isolation (Early Adulthood) & and appropriate choices, children develop a healthy sense of initiative in \\
leaf 2.2.7 Generativity vs. Stagnation (Middle Adulthood)& planning and undertaking activities If, instead, adults
discourage the\\
leaf 2.2.8 Integrity vs. Despair (Older Adulthood) & pursuit of independent activities or dismiss them as silly and \\
$\cdots$ & bothersome, children develop guilt about their needs and desires.\\
3 Chapter Three: Domains in Development\\
\midrule
\multicolumn{2}{l}{\textbf{Mistral}: 8.3.3 Moral Development, leaf 7.3.2 Hitting/Scratching and Temper Tantrums, \textbf{leaf 2.2.3 Initiative vs. Guilt (Preschool Years)}}\\
\multicolumn{2}{l}{\textbf{BM25}: 8.3.2 Moral Development}\\
\multicolumn{2}{l}{\textbf{DPR: }2.2.2 Autonomy vs. Shame/Doubt (Toddlerhood)}\\
\multicolumn{2}{l}{\textbf{DSI}: 2.2.2 Autonomy vs. Shame/Doubt (Preschool Years)}\\
\multicolumn{2}{l}{\textbf{{\sys}}: \textbf{2.2.3 Initiative vs. Guilt (Preschool Years)}}\\
\midrule
\bottomrule
\multicolumn{2}{c}{\textbf{Example 2:}} \\
\toprule
\multicolumn{2}{l}{\textbf{Book}: \href{https://open.umn.edu/opentextbooks/textbooks/1013}{Nursing Fundamentals}}\\
\multicolumn{2}{l}{\textbf{Query}: In what way does a chronic illness affect an elderly person's ability to perform daily activities?}\\
\midrule
\textbf{Table of Contents} & \textbf{Content}\\
$\cdots$\\
leaf 30.5 Spiritual Care of Self & Applying the Nursing Process\\
leaf 30.6 Putting It All Together& ...\\
31 Care of the Older Adult & It is also important to consider the impact of chronic disease\\
leaf 31.1 Care of the Older Adult Introduction & on
their ability to function and complete Activities of Daily Living\\
leaf 31.2 Basic Concepts &(ADLs). older adults who are able to perform ADLs without assistance\\
leaf \textbf{31.3 Applying the Nursing Process} & consider themselves healthy.\\
\midrule
\multicolumn{2}{l}{\textbf{Mistral}: \textbf{31.3 Applying the Nursing Process}}\\
\multicolumn{2}{l}{\textbf{BM25}: 11.2 Sensory Impairments Basic Concepts}\\
\multicolumn{2}{l}{\textbf{DPR: } 31.2 Basic Concepts}\\
\multicolumn{2}{l}{\textbf{DSI}: 31.2 Basic Concepts}\\
\multicolumn{2}{l}{\textbf{{\sys}}: \textbf{31.3 Applying the Nursing Process}}\\
\midrule
\bottomrule
\multicolumn{2}{c}{\textbf{Example 3:}} \\
\toprule
\multicolumn{2}{l}{\textbf{Book}: \href{https://open.umn.edu/opentextbooks/textbooks/1435}{Introduction to Comparative Government and Politics}}\\
\multicolumn{2}{l}{\textbf{Query}: What are irregular armed organizations and how are they used by states?}\\
\midrule
\textbf{Table of Contents} & \textbf{Content}\\
9 Chapter 5: Non-Democracies and Democratic Backsliding\\
$\cdots$ & Another powerful instrument of repression are paramilitaries.\\
9.2 Strategies for staying in power & These refer to groups with access 
to military-grade weapons and training\\
leaf 9.2.1 \textbf{Institutional channels}& yet they are not part of the national military.\\
leaf 9.2.2 Cultural and ideological controls & They are “irregular armed organizations that carry out acts of violence\\
9.3 Varieties of non-democracy & against civilians on behalf of a state,” \\
$\cdots$\\
\midrule
\multicolumn{2}{l}{\textbf{Mistral}: 9.3.5 Illiberal and hybrid regimes}\\
\multicolumn{2}{l}{\textbf{BM25}: 15.3.1 Insurgencies/Civil Wars}\\
\multicolumn{2}{l}{\textbf{DPR}: 15.2.2 External State-Sponsored Political Violence (State-Sponsored Terrorism)}\\
\multicolumn{2}{l}{\textbf{DSI}: 15.2.1 Internal State-Sponsored Political Violence (Government Terrorism)}\\
\multicolumn{2}{l}{\textbf{{\sys}}: \textbf{9.2.1 Institutional channels}}\\
\end{tabular}
}
\end{table*}

\begin{table*}[tb]
\centering
\small
\caption{Recall@1, Recall@3 and nDCG@3 for {\ben}; Haiku is Claude haiku 4.5}
\label{table:results_r1}
\resizebox{\textwidth}{!}{%
\begin{tabular}{lllllll|l}
\toprule
 & \textbf{Education} & \textbf{Finance} & \textbf{Law} & \textbf{Med} & \textbf{NatSci} & \textbf{SocSci} & \textbf{Avg}\\
 \midrule
Mistral & 14.8/17.8/16.6 & 16.4/19.4/18.2 & 13.6/15.7/14.9 & 13.7/17.0/15.7 & 13.1/15.3/14.4 & 10.9/13.4/12.4 & 13.8/16.4/15.4\\
BM25 & 58.7/75.8/68.8 & 55.0/76.8/67.8 & 62.0/77.7/71.3 & 59.6/78.7/70.8 & 58.7/78.1/70.1 & 62.8/78.3/72.0 & 59.5/77.6/70.1\\
Haiku & 59.8/80.4/72.0 & 37.7/56.3/48.4 & 41.1/60.6/52.3 & 46.1/65.0/57.2& 54.6/73.7/65.7 & 34.0/48.8/42.6 & 45.6/64.1/56.4 \\
DPR & 71.5/86.8/80.6 & 66.4/85.5/77.7 & 69.6/84.1/78.2 & 72.7/88.8/82.2 & 63.0/84.1/75.7 & 68.8/83.1/77.3 & 68.7/85.4/78.6\\
DSI & 76.2/84.1/80.9 & 78.1/87.8/83.9 & 73.3/82.9/79.0 & 82.1/89.9/86.8 & 76.2/84.7/81.3 & 75.4/82.2/79.4 & 76.9/85.3/81.9\\
{\sys} & \textbf{83.3}/\textbf{91.2}/\textbf{88.0} & \textbf{82.8}/\textbf{91.4}/\textbf{88.0} & \textbf{80.8}/\textbf{89.4}/\textbf{85.9} & \textbf{86.1}/\textbf{93.2}/\textbf{90.3} & \textbf{80.6}/\textbf{90.4}/\textbf{86.4} & \textbf{81.8}/\textbf{89.0}/\textbf{86.1} & \textbf{82.6}/\textbf{90.8}/\textbf{87.5}\\
\bottomrule
\end{tabular}
}
\end{table*}

Table~\ref{table:results_r1} compares {\sys} with all the baselines listed in Section~\ref{sec:baselines} and addresses the following two research questions:
\begin{itemize}
    \item \textbf{RQ1}: Does {\toc} based training makes {\sys} more accurate as a retriever?
    \item \textbf{RQ2}: Can we finetune an LLM to learn the new task of generating leaf nodes from {\toc}?
\end{itemize}

\noindent Our key findings are as follows:
\begin{enumerate}
    \item {\sys} outperforms strong baselines such as BM25, DPR and fine-tuned DSI
    \item R@1 for Mistral is much lower than BM25 -- this suggests the LLM not only needs to learn the new task of picking the best section but also needs to ingest knowledge from the corpus
    \item DPR performs much better than BM25 but worse than DSI. This result is expected, as NV-Embed-v2 is not fine-tuned for {\ben}
    \item Tuning with {\toc} as input helps {\sys}: The only difference in the input between DSI and {\sys} is that {\sys} leverages {\toc}. {\sys} outperforms DSI by $7.4\%$, as it is able to access the entire structure of the book and better align it with the queries during fine-tuning. DSI on the other hand must learn the semantic alignment between queries and sections by looking at all the training data. We can thus conclude that \textbf{RQ1} and \textbf{RQ2} are answered in affirmative.
\end{enumerate}
\textbf{Statistical significance testing} was done for the Recall@1 performance difference between DSI and {\sys}.  We follow the randomization test tailored to retrieval systems as described in ~\citet{sig_test}. The null hypothesis for the significance testing is that outputs from SystemA (DSI) and SystemB ({\sys}) may belong to the same underlying distribution and the difference of performance between SystemA (DSI) and SystemB ({\sys}) is because of the sampling variance. We test the null hypothesis with the traditional $p$ value of 0.05 at the domain level i.e. for each of the $6$ domains separately (by combining all the books from a single domain). The significance test results conclude that the difference in DSI and {\sys} is indeed statistically significant for all the $6$ domains where the null hypothesis is successfully invalidated. More specifically, through our sample runs for randomization test, we find that the probability of seeing this scale of Recall@1 performance difference as observed in Table~\ref{table:results_r1} when being randomly sampled from the same distribution falls below $p=0.05$. As with any statistical significance testing, the key takeaway from these experiments is a confirmation that the performance improvement seen in {\sys} is statistically significant and therefore, it is a property of the {\sys} design and not because of the data set size or data distribution in {\ben}.

\subsection{Ablations}
\label{subsec:ablation}
\begin{figure}[h]
\caption{Recall@1 for each leaf across {\ben}. {\sys} shows high accuracy even with a low number of training examples}
\includegraphics[width=\linewidth]{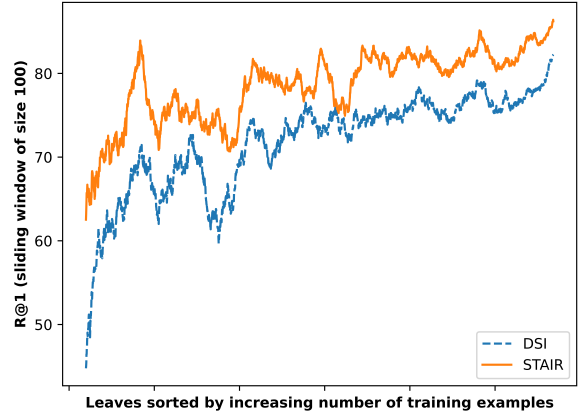}
\label{fig:ablation_b}
\end{figure}

We conduct ablation studies to study the impact of using {\toc} as an input. Two key conclusions are as follows:

\noindent\textbf{{\toc} reduces hallucinations}: We define hallucination as the generation of a non-leaf node (invalid document identifier). Figure ~\ref{fig:ablation_a} illustrates the number of hallucinations per leaf node. We observed that the hallucination rate of {\sys} remains nearly constant at close to zero, regardless of the number of training examples. This indicates that {\sys} has effectively learned to generate outputs solely from the leaf nodes, making it inherently less prone to hallucination. In contrast, DSI, which does not use {\toc}, exhibits a higher hallucination rate than {\sys}, particularly for leaf nodes with fewer training examples.\\
\textbf{Remembering the structure is hard without {\toc}} Figure~\ref{fig:ablation_b} shows that for leaf nodes with less number of training examples, the Recall@1 difference between DSI and {\sys} is much higher. This supports our hypothesis, that without {\toc}, DSI needs to discover the book structure and also remember it within its parameters. DSI attempts to learn this via the training data on query and associated {\toc} leaf node pairs. Thus, leaf nodes which have low representations in training data are prone to be missed by DSI. In contrast, {\sys} does not need to ``remember'' the leaf nodes. As the number of training examples for a leaf node increase, Recall@1 gap between DSI and {\sys} decreases -- although {\sys} consistently achieves higher Recall@1 numbers.

\subsection{Error Analysis with Anecdotal Examples}
We observe that out-of-the-box Mistral's error rate is 86.20\%, and  26.81\% of its mistakes are as it predicts a non-leaf node, while 23.86\% are hallucinations. This suggests that Mistral out-of-box needs knowledge ingestion and task finetuning. For DSI, the error rate drops to 24.31\%, with 3.25\% of its predictions being non-leaf nodes (suggesting knowledge infusion helps). {\sys} has the lowest error rate of 18.67\%, with only 0.05\% of its predictions being non-leaf nodes.

\begin{figure}[h]
\caption{As number of training examples decrease, hallucination rate for DSI increases}
\includegraphics[width=\linewidth]{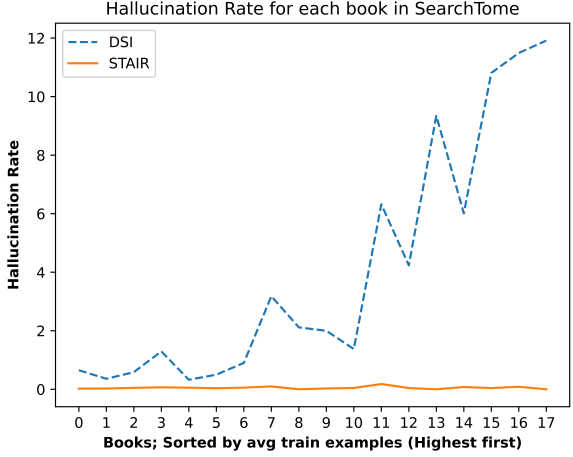}
\label{fig:ablation_a}

\end{figure}

\eat{
\section{Significance tests}

\begin{table}[!htb]
\begin{tabular}{lllll}
\toprule
\textbf{Domain}    & \textbf{Total} & \textbf{(a) TOC!= DSI} & \textbf{(b) ToC-DSI} & \textbf{(c)DSI-ToC} \\
\midrule
Finance   & 16415 & 3883      & 1857    & 1053    \\
\midrule
Education & 15326 & 4387      & 2075    & 1014    \\
\midrule
Law       & 27472 & 9076      & 3764    & 2017    \\
\midrule
Medicine  & 17346 & 3355      & 1709    & 865     \\
\midrule
Nat\_sci  & 15719 & 4335      & 1860    & 1327    \\
\midrule
Soc\_sci  & 10971 & 2610      & 1209    & 487    \\
\bottomrule
\end{tabular}
\label{significance_testing}
\caption{Statistics on total number of instances per domain and number of instances where (a) DSI and ToC output differ (b) TOC is correct and DSI is incorrect (c) DSI is correct and ToC is incorrect  }
\end{table}
}
Table~\ref{table:pred_examples} lists a few qualitative examples. We discuss next the first example ``How do preschoolers react when caregivers and teachers belittle their autonomous actions'' from the book ``The Whole Child: Development in the Early Years'':

\textbf{Mistral} could correctly guess that the query can be answered by ``2.2.3 Initiative vs. Guilt (Preschool Years)'' using its world knowledge. It doesn't understand that it needs to pick only one option -- due to lack of the task training\\
\textbf{BM25} suffers from the known bias of keyword matching where and picks the section ``8.3.2. Moral Development'' purely because of highest lexical overlap with query tokens, which is the incorrect answer\\
\textbf{DPR} is used out of the box and does a word sensed matching where it predicts ``2.2.2 Autonomy vs. Shame/Doubt (Toddlerhood)'' being most relevant to the query tokens mentioning ``belittling of autonomous actions''. However, without the corpus specific training it falls short of identifying the correct section.\\
\textbf{DSI} is finetuned for the corpus but faulters to find the precise section. It errs just like DPR in this example but might have chosen a more sensible match ``2.2.2 Autonomy vs. Shame/Doubt (Preschool years)'' which takes into account the importance of keyword ``preschool'' for getting a better topic match. However, careful consideration shows this is a hallucinated header not present in the input {\toc}. This correlates with our ablation in figure~\ref{fig:ablation_a} on DSI hallucination.\\
\textbf{\sys}: Our proposed system {\sys} correctly identifies the gold leaf node as ``Initiative vs. Guilt (Preschool Years)''. 

The comparison between DPR, DSI and {\sys} is interesting as they all are built on top odf the same base LLM -- Mistral. The difference between DSI and {\sys} largely points towards the importance of having {\toc} as an input during training, which makes it easier for the LLM to align much better to the corpus. Other anecdotal examples in Table~\ref{table:pred_examples} also enumerate the same partial ordering seen among the baselines and provide more qualitative analysis to support the claim that the design of {\sys} helps it do a much better retrieval.

\section{Conclusion}
\label{sec:conclusion}
We introduce {\sys}, a novel LLM based IR system that leverages {\toc} to store and retrieve information from its parameters. Given a query, {\sys} generates the most probably leaf section header from a {\toc} which could answer it. We use {\basellm} to demonstrate that it is possible to instruction finetune the modern day LLMs to efficiently use the {\toc} structure to generate the correct leaf node while reducing the hallucinations to almost zero. {\sys} outperforms all strong baselines such as BM25, finetuned DSI and DPR and achieves a Recall@1 score of 82.6\% which is around $7.4\%$ gain over the next best system (DSI). We release a new comprehensive benchmark {\ben} across $6$ domains with $18$ books comprising of train, dev and test splits to further research in this novel direction.

As a future work, we want to explore directions where a {\toc} like structure is created dynamically over an unseen search corpus, as has been proposed by some previous works. We envision that having a {\toc}-based retrieval paradigm will gain more traction in future for agentic frameworks needing multi-hop retrieval and reasoning over retrieved context. To that end, we want to develop {\sys} to work in a complete zero shot setup where it iteratively retrieves {\toc} leaf nodes and makes intelligent decisions by reasoning on the content of the leaf node to do precise information retrieval. 

\section{Limitations}
Our current evaluation is limited to corpora where a global structure exists. While this setting is suitable for initial validation, it may not fully represent the diversity of real-world use cases. In future work, we plan to extend our evaluation to standard benchmarks by artificially inducing a Table of Contents structure. Additionally, we aim to test our model on enterprise datasets where such a structure already exists at a very large scale (possibly millions of URLs in the corpus).


\bibliography{emnlp}




\end{document}